\documentclass[journal]{IEEEtran}

\usepackage{amsmath,amsfonts}
\usepackage{algorithmic}
\usepackage{algorithm}
\usepackage{array}
\usepackage[caption=false,font=normalsize,labelfont=sf,textfont=sf]{subfig}
\usepackage{textcomp}
\usepackage{stfloats}
\usepackage{url}
\usepackage{verbatim}
\usepackage{graphicx}
\usepackage{cite}
\usepackage{multirow}
\usepackage{xcolor}
\usepackage{ulem} 
\usepackage[hidelinks]{hyperref} 
\usepackage{orcidlink}

\begin{document}

\title{Robustness-enhanced Myoelectric Control with GAN-based Open-set Recognition}

\author{

    \IEEEauthorblockN{
        Cheng Wang\orcidlink{0000-0001-7527-9702}, 
        Ziyang Feng\orcidlink{0009-0006-4404-1527}, 
        Pin Zhang,~ \\
        Manjiang Cao, 
        Yiming Yuan, 
        Tengfei Chang\orcidlink{0000-0001-9589-7794}
    }
    
    \thanks{
        Cheng Wang, Manjiang Cao, Yiming Yuan and Tengfei Chang are with the 
        IoT Thrust, Information Hub, The Hong Kong University of Science and Technology (Guangzhou), 
        Guangdong, 511400, China. (e-mail: cwang199@connect.hkust-gz.edu.cn; yyuan861@connect.hkust-gz.edu.cn; mcao999@connect.hkust-gz.edu.cn; tengfeichang@hkust-gz.edu.cn).
        \textit{(Corresponding author: Tengfei Chang).}
        
        Ziyang Feng and Pin Zhang are with the School of Automation \& Electrical Engineering, University of Science and Technology Beijing, Beijing 100083, China. (e-mail: fengziyang13@126.com; m202110531@xs.ustb.edu.cn).
    }
}

\maketitle

\begin{abstract}

Electromyography (EMG) signals are widely used in human motion recognition and medical rehabilitation, yet their variability and susceptibility to noise significantly limit the reliability of myoelectric control systems.
Existing recognition algorithms often fail to handle unfamiliar actions effectively, leading to system instability and errors.
This paper proposes a novel framework based on Generative Adversarial Networks (GANs) to enhance the robustness and usability of myoelectric control systems by enabling open-set recognition.
The method incorporates a GAN-based discriminator to identify and reject unknown actions, maintaining system stability by preventing misclassifications.
Experimental evaluations on publicly available and self-collected datasets demonstrate a recognition accuracy of 97.6\% for known actions and a 23.6\% improvement in Active Error Rate (AER) after rejecting unknown actions.
The proposed approach is computationally efficient and suitable for deployment on edge devices, making it practical for real-world applications.

\end{abstract}


\section{Introduction}


\IEEEPARstart{M}{odern} myoelectric control systems are designed to interpret movement intentions from electromyography (EMG) signals~\cite{allard2020interpreting}, requiring reliable methods for EMG signal pattern recognition. 
These systems process EMG signals to extract information about muscle activity. 
Compared to manual or semi-automated prosthetics and exoskeletons, using EMG to decode neural signals enables users to control these devices more intuitively and naturally~\cite{ahkami2023electromyography,fleming2021direct}. 
Nonetheless, even the most advanced commercial myoelectric control devices require a period of adaptation and training to achieve effective prosthetic control~\cite{fleming2021direct,gu2021asn}. 
Moreover, their recognition algorithms must be tailored to each user to ensure optimal performance. 
The diversity of real-world scenarios and environments poses significant challenges to the stability and accuracy of myoelectric control systems~\cite{ahkami2023electromyography,ketyk2019domain}.


This instability arises due to two main factors: 
    the variability in signal characteristics collected from different individuals, and 
    environmental errors during the signal acquisition process. 
The variability in signal characteristics is influenced by factors such as muscle fatigue and individual muscle properties, leading to variations in EMG signal outputs for the same movement. 
Environmental errors include challenges such as electrode displacement and changes in skin impedance. 
For a specific individual, the impact of environmental errors is generally less significant than that of signal variability over short time periods. 
According to Atzori et al.~\cite{atzori2016deep}, an average accuracy of only 66\% is achieved when testing involves multiple repetitions of signals from the same subject performing successive repetitive actions. 
This phenomenon, referred to as ``inter-subject/session'' variability, is further discussed in ~\cite{xiong2021deep}, highlighting the differences in EMG signals across individuals and over time. 
Such low recognition accuracy significantly contributes to the instability of the system.


Improving the methods for EMG signal collection or enhancing the performance of pattern recognition algorithms can help reduce the effects of this instability. 
Research on improving collection methods~\cite{wang2021towards} is limited and predominantly focuses on non-surface EMG (sEMG) signals, such as intramuscular EMG (iEMG)~\cite{atzori2014classification,zhang2019cooperative}. 
Currently, most research efforts are directed toward developing more robust pattern recognition algorithms, with a variety of machine learning and deep learning techniques being explored in the field of EMG signal pattern recognition.

In addition to refining algorithms, further research has focused on signal pre-processing techniques within deep learning methods. 
Some studies employ specialized approaches, such as transfer learning, for EMG signal recognition, showing promising results in scenarios with limited inter-session~\cite{du2017surface} and inter-subject recognition~\cite{zhang2019cooperative,ctallard2019deep,allard2020interpreting,ctAllard2020unsupervised,demir2019surface}. 
In these cases, each individual and time session is treated as a distinct domain, framing the problem as a multi-domain learning challenge~\cite{ctallard2019deep}. 
These machine learning and deep learning strategies have significantly improved the robustness of EMG signal pattern recognition.



This paper introduces an enhanced deep learning approach leveraging Generative Adversarial Networks (GANs) to improve the stability of myoelectric control systems. 
The proposed method involves performing experiments with non-target patterns, treating the corresponding EMG signals as open sets to distinguish between correct and false classifications. 
Initially, a simple Convolutional Neural Network (CNN) model is utilized to classify K-patterns derived from known EMG gestures. 
Based on the K-dimensional prediction output of the CNN model, an open-set discriminator is employed to identify known classifications while rejecting unknown ones. 
When the output is applied to an actuator, such as a prosthetic or exoskeleton, only recognized classifications are executed, while unknown actions are disregarded to prevent operational errors. 
This open-set discriminator model is particularly efficient due to its simplicity, minimizing the computational burden on hardware devices.

This article is structured as follows. 
Sec.~\ref{sec:background} details the related classification methods applicable to gesture recognition using EMG signals. 
Sec.~\ref{sec:gan-openset} describes the principles of GAN-based Open-Set Recognition. 
Sec.~\ref{sec:enhanced} introduces our enhanced EMG gesture recognition approach utilizing the GAN-based Open-Set Recognition technique. 
Sec.~\ref{sec:exp} explains the experiments conducted to evaluate our method, including three sub-experiments that progressively extend the applicability of this approach. 
Sec.~\ref{sec:eval} analyzes the experimental results in detail. 
Sec.~\ref{sec:discussion} further explores the relationship between the proposed method, confidence-based rejection, and transfer learning. 
Finally, Sec.~\ref{sec:conc} concludes this article.

\section{Background}
\label{sec:background}


Multiple deep learning-based approaches have been explored to enhance pattern recognition accuracy. 
In \cite{ctallard2019deep}, Côté-Allard et al. demonstrate that applying Complex Wavelet Transform (CWT) to CNN training data for gesture classification significantly improves recognition accuracy compared to using raw data directly. 
Subsequently, the authors propose new methods to further enhance accuracy by employing transfer learning. 
In transfer learning, the domain is defined as samples collected from different individuals and time sessions, framing the task as a multi-domain problem~\cite{allard2020interpreting}. 
Transfer learning has been shown to improve EMG signal recognition performance, particularly in ``inter-subject/session'' scenarios~\cite{ctallard2019deep,allard2020interpreting,ctAllard2020unsupervised,demir2019surface}.


Beyond accuracy, the active error rate (AER) of EMG signals is one of the primary factors limiting the adoption of myoelectric control systems in clinical practice~\cite{hargrove2007areal}. 
AER refers to the rate at which the device misinterprets muscle movement patterns, leading to erroneous actions. 
Hargrove et al. argue that halting an incorrect action costs less than executing it, as correcting an error is often more time-consuming~\cite{hargrove2007areal}. 
If incorrectly identified results can be detected and subsequently rejected by the actuator—i.e., maintaining immobility or a default state—unnecessary overhead can be significantly reduced. 
This would decrease operational costs and improve the clinical "usability" of these systems.


Many studies in this area focus on a key approach: increasing confidence in the algorithm's output. 
Confidence refers to the reliability of the model's classification results. 
Only high-confidence gesture classifications are executed, while low-confidence outputs are blocked, resulting in an idle state. 
Scheme et al.~\cite{scheme2011selective} proposed a novel ``1 vs. all'' method to compare the Active Error Rate (AER) and Total Error Rate (TER, which includes both active and idle states) across several machine learning methods. 
Subsequently, Scheme et al.~\cite{scheme2013confidence} introduced a rejection strategy following linear discriminant analysis (LDA) and defined multiple performance measures to evaluate the effectiveness of EMG control systems. 
Robertson et al.~\cite{robertson2019effects} applied these measures to study the impact of rejection thresholds on different error metrics. 

At early stage, the selection of confidence levels in EMG signal recognition primarily relied on machine learning, with manually preset parameters such as rejection thresholds~\cite{scheme2011selective}. 
With the advancement of deep learning in recent years, adaptive algorithms~\cite{ketyk2019domain} and transfer learning methods~\cite{ctAllard2020unsupervised,demir2019surface} have been introduced to improve model accuracy, although less attention has been given to the stability of myoelectric control systems. 
For example, Bao et al.~\cite{bao2022cnn} proposed a method that multiplies the CNN prediction output by a matrix and maps the results to a step function to compute the confidence score. 
This approach outperformed methods based on maximum posterior probability and inverse entropy. 

The work on confidence estimation for myoelectric control predominantly focuses on known data. 
Real-world scenarios, where prosthetic users are unlikely to restrict their movements to predefined classes, remain largely underexplored. 
The literature offers limited discussion on strategies to improve the "usability" of myoelectric control systems when faced with unknown movement categories.

\begin{figure}[!t]
	\centering
	\includegraphics[width=0.8\columnwidth]{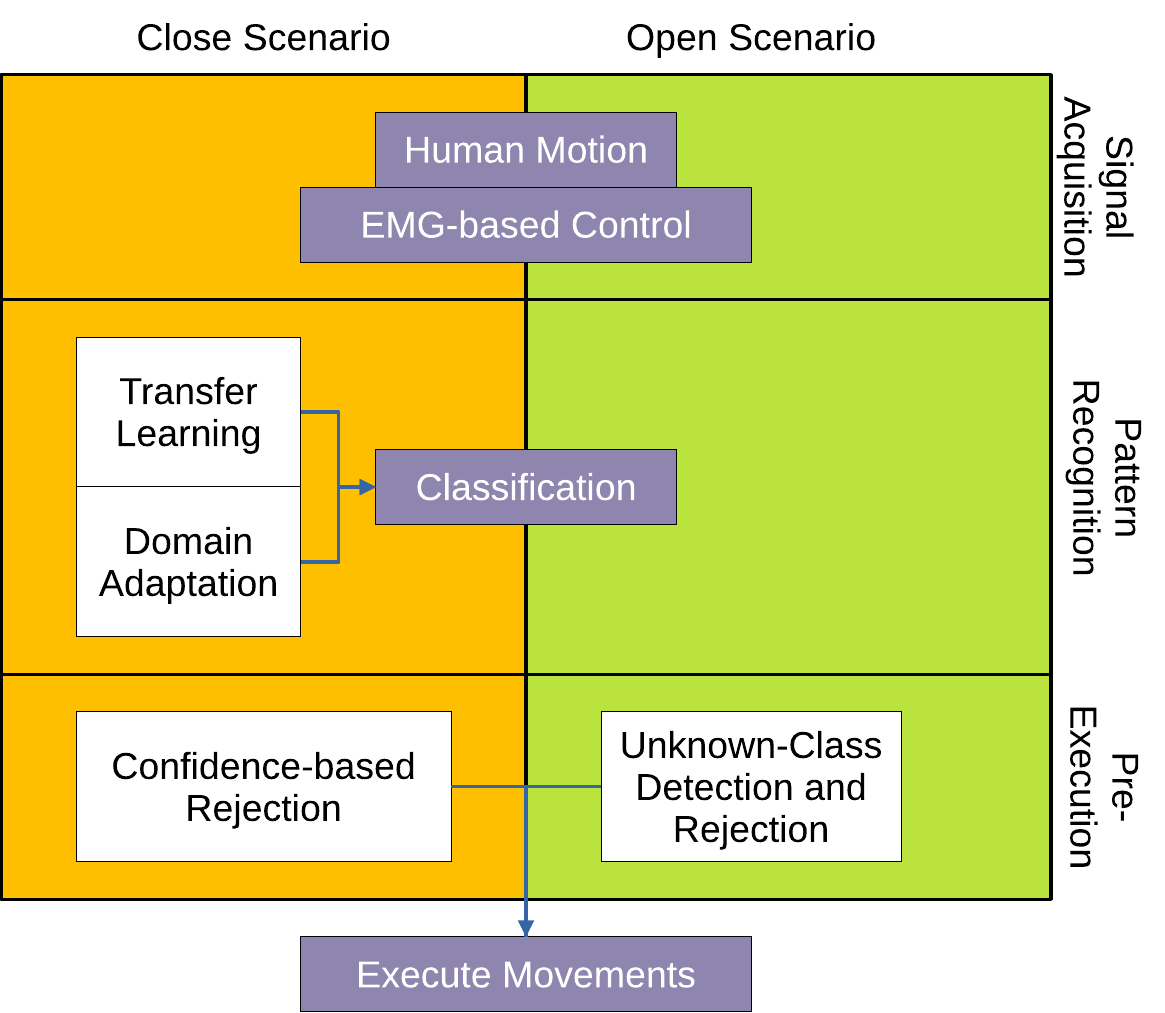}
	\caption{
            Three phases of EMG-based signal control flow for human motion detection: signal acquisition, pattern recognition and pre-execution.
            The techniques/approaches focus on known classes only are categorized as \textit{Close} scenario, while the ones focusing on both known and unknown classes are categorized as \textit{Open} scenario.
            The pre-execution phase involves explorations like confidence-based rejection.
            In \textit{Open} scenario, the focus shifts to unknown class detection and rejection.
        }
	\label{fig:emg_control_flow}
\end{figure}

Another issue is that the most EMG pattern recognition studies primarily focused on existing sample classes within the training set, yielding positive effects on enhancing classifier confidence, accuracy, and robustness. 
However, the identification of unrecognized categories resulting from inter-subject/session variability remains largely unexplored. 
In this paper, we propose an approach that extracts a subset of samples from a specific class within the collected data and designates them as an ``unknown class." 
The model is then exclusively trained on the remaining ``known classes." 

We employ a GAN discriminator as the core model. 
During the GAN training process, the discriminator learns to differentiate between known and unknown classes, enabling the detection of unknown classes in scenarios involving previously unseen samples. 
By integrating this discriminator with a CNN classifier, the model effectively classifies known classes while rejecting instances of unknown classes. 
This approach has been demonstrated to significantly enhance the confidence and robustness of the EMG signal classifier in scenarios involving unknown signals.
The outcome of proposed approach is applied at the pre-execution phase of the EMG-based human motion detection system as shown in Fig.~\ref{fig:emg_control_flow}.

The contribution of this paper is in three folders: 
\begin{itemize}
    \item We proposed a novel EMG signal recognition approach using a GAN-based Open-Set technique, significantly enhancing the robustness of Myoelectric Control Systems.
    \item We evaluated the performance of the proposed method using the Ninapro DB1 dataset. 
    The results demonstrated the effectiveness of the proposed approach compared to both closed-set methods and open-set methods without GAN, across various aspects.
    \item We conducted experiments with real hardware and collected real EMG signal samples, utilizing the trained discriminator to reject unknown classes. 
    The results showed a 23.6\% improvement with the discriminator and a recognition accuracy of 97.6\%.
\end{itemize}

\section{GAN-based Open-Set Recognition}
\label{sec:gan-openset}

Most recognition models are trained on a fixed number of predefined classes within a dataset. 
When data of an unknown type is input into the model, it is often misclassified as one of the known classes, which can lead to malfunctions, such as unintended prosthetic limb movements. 
To address this challenge, open-set recognition models have been developed~\cite{zhou2022learning}. 
These models introduce specially prepared outlier data as unknown classes~\cite{ge2017generative} and utilize a GAN discriminator to identify and reject such unknown data~\cite{schlegl2017unsupervised}. 
Furthermore, a GAN generator can be employed to produce synthetic outlier data, which is subsequently used to train the discriminator~\cite{neal2018open}.


The Generative Adversarial Network (GAN) was originally introduced for generating realistic images~\cite{goodfellow2014generative}. 
A GAN is composed of two networks: the generator $G$ and the discriminator $D$. 
The discriminator is trained to distinguish real samples from fake ones, while the generator is trained to create synthetic samples that can fool the discriminator, thereby increasing its error rate. 
In this study, the generator is implemented as a convolutional neural network, whereas the discriminator is a fully connected neural network.

GANs evaluate the performance of the discriminator using cross-entropy loss, defined as $L_D = -\sum_{i=1}^{m} P_i \log(Q_i)$, where $P_i$ represents the true probability distribution and $Q_i$ denotes the predicted probability distribution. 
Since there are only two mutually exclusive states, \texttt{known} and \texttt{unknown}, $P_i$ takes a value of either 1 or 0. 

The loss for real samples and synthesized samples, generated by the generator, can be represented as $L_D^{R} = -\log(D(x_i))$ and $L_D^{G} = -\log(1 - D(G(z_i)))$, respectively. 
$x_i$ represents a real sample, and $G(z_i)$ denotes a synthetic sample generated from random noise. 
To ensure a consistent gradient direction, the loss for synthesized samples is expressed as $-\log(1 - D(G(z_i)))$ instead of $-\log(D(G(z_i)))$, as the discriminator aims to accept real samples while rejecting synthetic ones. 
The final loss function, $V_D$, for updating the discriminator is calculated as the average of the losses for both real and synthesized samples, as illustrated in Eq.~\ref{eq:gan_vd}.

\begin{equation}
    V_D = \frac{1}{m}\sum_{i=1}^{m}[logD(x_i)+log(1-D(G(z_i)))]
    \label{eq:gan_vd}
\end{equation}

Since the gradient update for the generator also depends on the discriminator's output, the two networks are cascaded during the generator's update process. 
As a result, the update formula $V_G$ for the generator only includes the portion of $V_D$ that involves $G(z)$, as shown in Eq.~\ref{eq:gan_vg}.

\begin{equation}
    V_G =\frac{1}{m}\sum_{i=1}^{m}[log(1-D(G(z_i)))]
    \label{eq:gan_vg}
\end{equation}

In practice, $V_G$ represents a subset of $V_D$, and the complete equation is expressed as $V(D,G)$. 
The optimization objective for the discriminator is to maximize its ability to correctly distinguish real data, which corresponds to maximizing $V(D,G)$, as shown in Eq.~\ref{eq:gan_obj_d}.

\begin{equation}
    D_G^* = argmax_D V(D,G)
    \label{eq:gan_obj_d}
\end{equation}

Meanwhile, the objective for the generator, as shown in Eq.~\ref{eq:gan_obj_g}, is to minimize the second part while keeping $D_G^*$ constant.

\begin{equation}
    G^* = argmin_G V(G,D_G^*)
    \label{eq:gan_obj_g}
\end{equation}

By unifying the optimization objectives of the discriminator and the generator, and substituting the mean with the mathematical expectation, the classic formula for the GAN optimization objective is derived, as shown in Eq.~\ref{eq:deqn_ex1a}.

\begin{multline}
	\label{eq:deqn_ex1a}
	min_Gmax_DV(D,G) = E_{x-P_{data(x)}}[logD(x)]+\\
	E_{x-P(z)}[log(1-D(G(z)))]
\end{multline}

\section{Enhanced-Myoelectronic Control with GAN-based Open-set Recognition}
\label{sec:enhanced}

\begin{figure}[!t]
    \centering
    \includegraphics[width=\columnwidth]{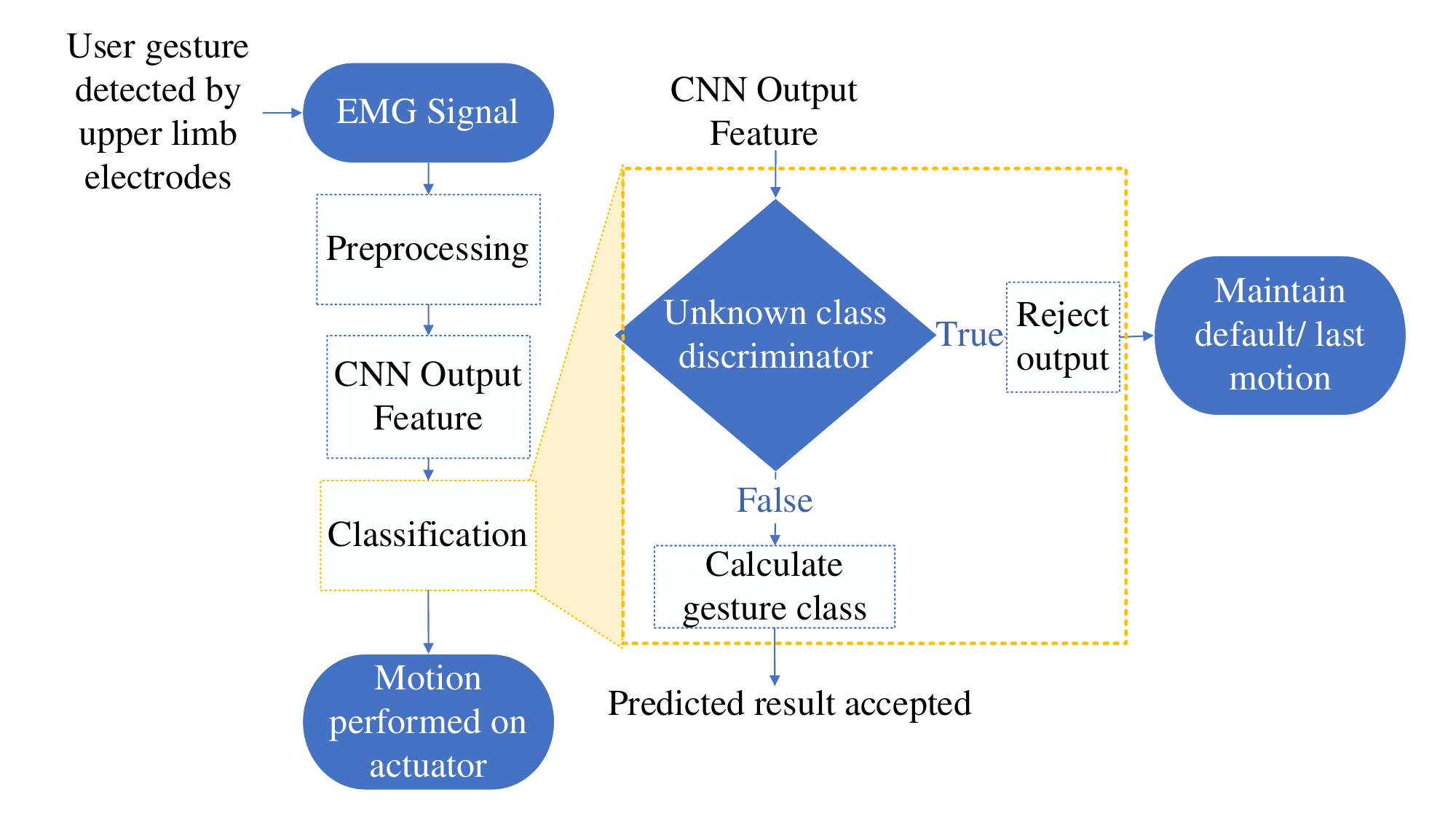}
        \caption{
            The Framework for GAN-based EMG Signal Pattern Recognition.
        }
    \label{fig:method_framework}
\end{figure}
 
The proposed framework using GAN-based Open-Set recognition to enhance Myoelectronic control is illustrated in Fig.~\ref{fig:method_framework}.
It contains three steps: data pre-processing, CNN feature extraction and classification in accepted actions, which are detailed in the section below respectively.

\subsection{Step 1: Data Pre-Processing}

During the data pre-processing phase, surface EMG signals are sampled simultaneously using 10 electrodes at a sampling rate of 1000~Hz. 
The collected samples are grouped into segments of 200~ms. 
Each segment is then fed into the CNN classifier for further processing. 

10 different gestures are performed during the sampling process, resulting in 10 distinct classes of samples. 
A subset of these classes is arbitrarily selected and labeled as \texttt{unknown}, while the remaining classes are labeled as \texttt{known}. 
The samples from the \texttt{known} classes are used for training the CNN and GAN models. 

\begin{figure}[!t]
	\centering
	\includegraphics[width=\columnwidth]{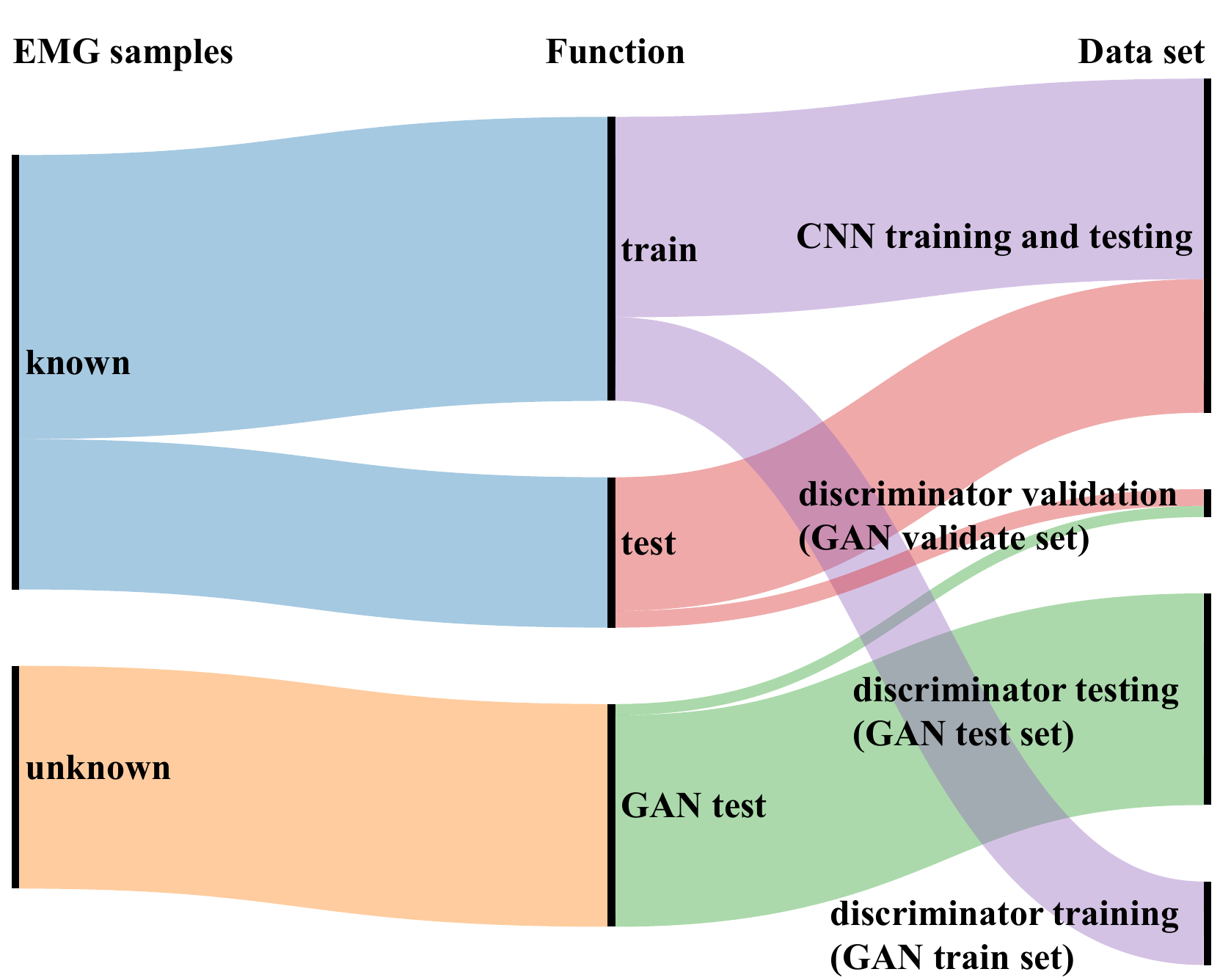}
	\caption{
            \texttt{known} and \texttt{unknown} class, test and train set, GAN and CNN training and testing.
        }
	\label{fig:sankey_dataset}
\end{figure}

To validate the trained \texttt{unknown} class discriminator, a certain portion of samples from both \texttt{known} and \texttt{unknown} classes is reserved for testing.
How all EMG samples are spited into \texttt{known} and \texttt{unknown} dataset, later being distributed for training, validating or testing by CNN-classifier and discriminator are illustrated in Fig.~\ref{fig:sankey_dataset}.

\subsection{Step 2: Calculate Model Feature Vector with CNN}

\begin{figure}[!t]
	\centering
	\includegraphics[width=\columnwidth]{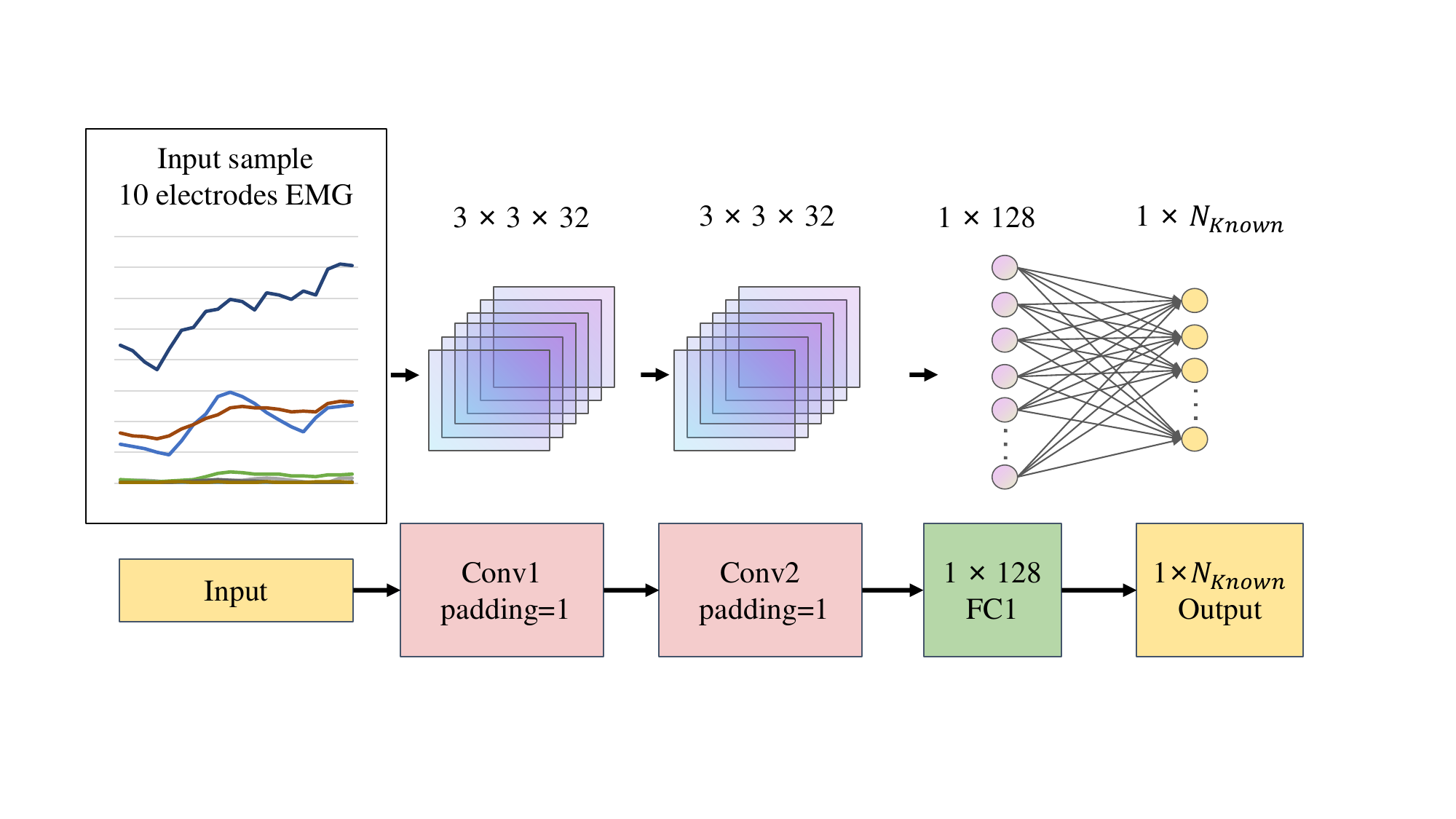}
	\caption{
        The architecture of CNN-based classifier. 
        The input to the CNN is an $N_{Channel} \times N_{Sample~points}$ matrix representing the EMG signals. 
        The input is processed through two convolutional layers, \textit{Conv1} and \textit{Conv2}, each comprising 32 $3\times3$ convolutional kernels. 
        Padding is set to 1 to maintain consistent input and output dimensions for each convolutional layer. 
        The output of \textit{Conv2} is flattened and passed to the fully connected layer, \textit{FC1}. 
        The number of neurons in the final layer is set to match the number of \texttt{known} classes, $N_{known}$.
    }
    \label{fig:cnn_arch}
\end{figure}

A CNN classifier is developed to classify \texttt{known} class samples. 
The architecture consists of two convolutional layers and two fully connected layers, as illustrated in Fig.~\ref{fig:cnn_arch}. 
The input to the CNN is an $N_{Channel} \times N_{Sample~points}$ matrix representing the EMG signals. 
$N_{Channel}$ is set to 10, corresponding to the number of electrodes, and $N_{Sample~points}$ represents the number of samples within a 200~ms time window. 

The input data is processed through two convolutional layers, \textit{Conv1} and \textit{Conv2}, each utilizing 32 $3\times{3}$ convolutional kernels. 
Padding is set to 1 to ensure consistent input and output dimensions across each convolutional layer. 
The output of \textit{Conv2} is flattened and passed to the fully connected layer, \textit{FC1}. 

The number of neurons in the output layer is set to match the number of \texttt{known} classes, $N_{known}$. 
Given that the total number of gesture classes does not exceed 50, the number of neurons in \textit{FC1} is chosen to be larger than the output layer, following an exponent scale of 2. 
As a result, $2^7 = 128$ neurons are selected for \textit{FC1}.

\subsection{Step 3: Classification with Trained Discriminator}

The goal of this step is to train the discriminator to identify \texttt{unknown} class samples and reject them, thereby preventing false actions. 
This process requires both \texttt{known} and \texttt{unknown} samples for training the discriminator.
Since the \texttt{unknown} EMG signal samples collected in the first step are preserved for the final evaluation of the discriminator, the generator in the GAN is used to produce synthetic \texttt{unknown} samples for training. 
The \texttt{known} EMG signal samples are directly utilized for discriminator training.

\begin{figure}[!t]
	\centering
	\includegraphics[width=\columnwidth]{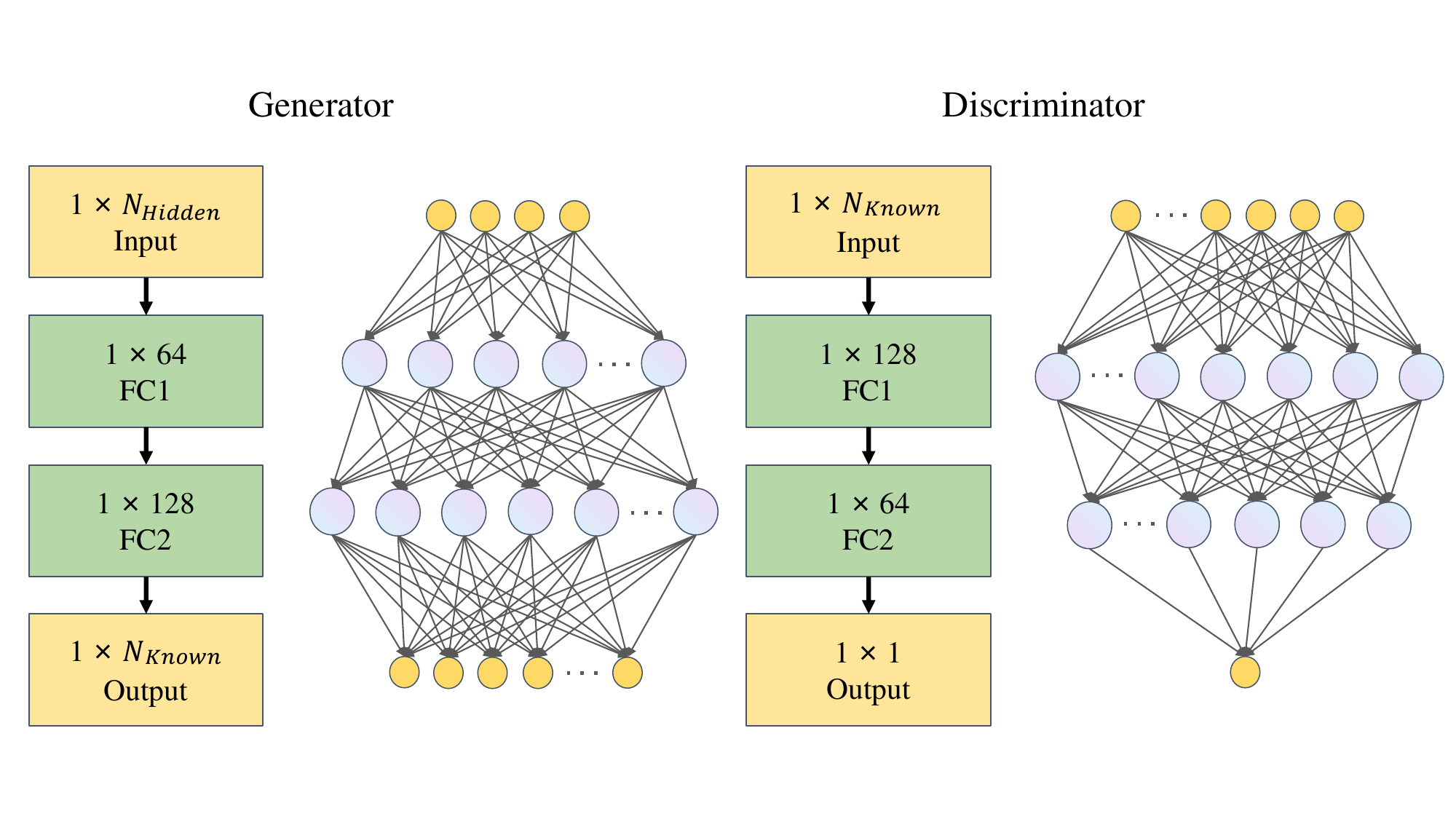}
	\caption{
            The GAN consists of two competing networks during the training process: the generator and the discriminator. 
            The generator takes random Gaussian noise of size $1 \times N_{Hidden}$ as input and produces an output vector of length $N_{Known}$. 
            The input to the discriminator has a size of $(1 \times N_{Known})$, and its output is a single numerical value between 0 and 1, representing the classification judgment. 
            The fully connected layers \textit{FC1} and \textit{FC2} in the discriminator consist of 128 and 64 neurons, respectively.
        }
	\label{fig:gan_arch}
\end{figure}

As shown in Fig.~\ref{fig:gan_arch}, the GAN consists of two competing networks during the training process: the generator and the discriminator. 
The generator takes random Gaussian noise of size $1 \times N_{Hidden}$ as input and outputs a vector of length $N_{Known}$. 
For effective generator training, the dimension of $N_{Hidden}$ is typically set smaller than $N_{Known}$. 
The discriminator's goal is to determine whether its input originates from real EMG samples or synthetic data generated by the generator. 
The discriminator's input is of size $(1 \times N_{Known})$, and it outputs a single numerical value between 0 and 1 as its judgment. 
The fully connected layers, \textit{FC1} and \textit{FC2}, in the discriminator have 128 and 64 neurons, respectively.

The discriminator's output represents the predicted probability of the input sample being real. 
A threshold value is required to determine whether the input sample is real or synthesized. 
This threshold is determined using a metric called the Area Under the Curve (AUC).

\begin{figure}[!t]
	\centering
	\includegraphics[width=\columnwidth]{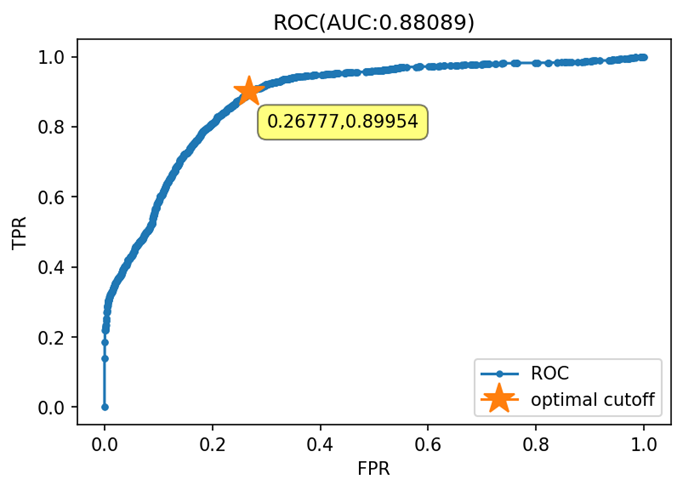}
	\caption{
        The ROC curve and its optimal cutoff are determined using the test set derived from self-collected data.
        The ROC curve depicts the performance of the binary classification model across a range of threshold settings.
        The True Positive Rate (TPR) represents the proportion of actual positive cases correctly identified by the model.
        The False Positive Rate (FPR) represents the proportion of actual negative cases incorrectly classified as positive by the model.
        The point closest to the top-left corner of the ROC curve corresponds to the optimal threshold value.
        This threshold minimizes the FPR while maximizing the TPR, ensuring an optimal trade-off between sensitivity and specificity.
        }
	\label{fig:roc_example}
\end{figure}

The AUC quantifies the entire two-dimensional area beneath the Receiver Operating Characteristic (ROC) curve, as shown in Fig.~\ref{fig:roc_example}. 
The ROC curve illustrates the trade-off between sensitivity (True Positive Rate, TPR) and specificity (False Positive Rate, FPR) as the classification threshold varies. 
The AUC provides a single scalar value between 0 and 1, representing the model's ability to discriminate between positive and negative classes. 
A higher AUC value indicates better performance by the discriminator. 
During the training process, the discriminator with the highest AUC value is selected as the \texttt{unknown} class discriminator.

The most upper-left corner point on the ROC curve of the selected discriminator is chosen as the threshold value. 
If the predicted probability produced by the discriminator is equal to or greater than this threshold, the input sample is classified as a \texttt{known} class and accepted. 
In this case, the final action or gesture corresponding to the class of the sample is executed. 
If the probability is lower than the threshold, the sample is rejected, and the system maintains its default or previous state until the next accepted sample is identified.

\section{Experiments}
\label{sec:exp}

This section details the data collection process for training the model, utilizing both practical experiments and publicly available datasets.
The collected data is subsequently used to determine the thresholds that enable the model to distinguish between \texttt{known} and \texttt{unknown} classes.

\subsection{Data Collection}

\subsubsection{Self-Collected Data}

\begin{figure}[!t]
	\centering
	\includegraphics[width=\columnwidth]{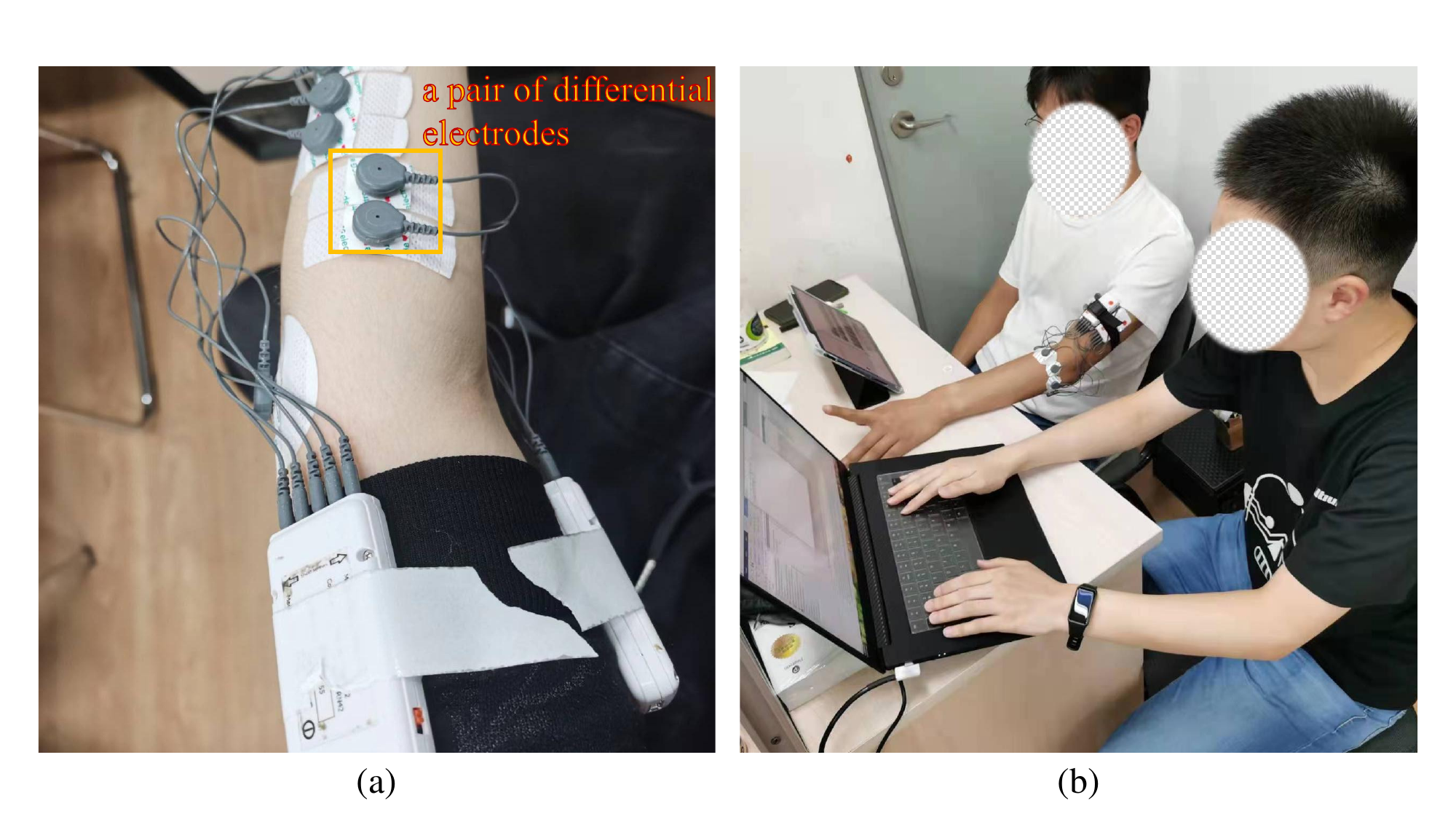}
	\caption{
        The gesture capture process begins with configuring the device to operate in differential mode, which enhances the signal-to-noise ratio.
        Differential electrodes are carefully positioned in close proximity, as indicated by the yellow box in Fig.~\ref{fig:gesture_collecting_process}(a).
        The Shimmer device collects EMG signals and transmits them to a computer in real-time via Bluetooth communication.
        During the recording session, the subject is guided to alternate between rest and movement states, as illustrated in Fig.~\ref{fig:gesture_collecting_process}(b).
        }
	\label{fig:gesture_collecting_process}
\end{figure}

This study was approved by The Hong Kong University of Science and Technology (Guangzhou) Committee on Research Practices ( HKUST(GZ)-HSP-2025-0125).
A commercial EMG signal acquisition device~\footnote{https://shimmersensing.com/product/shimmer3-gsr-unit/} was utilized for data collection.
The device includes five probes: four for signal acquisition and one for calibration.
In differential mode, the four acquisition probes are arranged into two pairs, with each pair positioned closely together, as shown in Fig.~\ref{fig:gesture_collecting_process}, to form a single channel and reduce common-mode error.
Each Shimmer3 device provides two channels for sampling EMG signals, allowing the simultaneous use of three devices to capture six channels.
The calibration probe is attached to the prominent part of the arm joint to mitigate DC bias.
EMG signals are transmitted to a computer in real time via Bluetooth communication.
The signal sampling rate is set to 1000 Hz to ensure high-fidelity signal acquisition while maintaining the quality of Bluetooth communication. 
All participants involved in the data collection process provided written informed consent prior to their participation in the study. All data were anonymized to protect participant privacy.

\subsubsection{Public Dataset}

The Ninapro public dataset~\cite{atzori2014electromyography} is the largest publicly available EMG dataset and serves as a widely used resource in EMG signal-related research.
The dataset is organized into ten categories, labeled from \textit{DB1} to \textit{DB10}.
For this study, we exclusively utilized data from the \textit{DB1} category.
The \textit{DB1} dataset contains over 50 distinct gesture categories of EMG signal samples.
During data recording, each gesture is performed for 10 seconds and repeated 10 times, with a 10-second rest period between repetitions.
Participants are seated with their arms resting on a table to reduce the influence of arm and body movements.
EMG signals are sampled at intervals of 50~ms.
Samples from each 200~ms time window are grouped together to create a format compatible with neural network processing.
This methodology produces 50 groups of samples for each 10-second recording period per gesture.
The $2^{nd}$, $5^{th}$, and $7^{th}$ groups of samples from each gesture are used as the validation set to evaluate the final trained model.

\subsection{Model Training}

The model training process adheres to the methodology outlined in Sec.~\ref{sec:enhanced}.
All deep learning code is implemented using PyTorch and is publicly available online~\footnote{github.com/cwdao/paper\_code\_robustness\_enhanced\_myoelectric\_control/}.
The training process is conducted on RTX 2080 GPUs.
To optimize the performance of the discriminator, a threshold is determined based on the AUC value produced by the discriminator.
As shown in Fig.~\ref{fig:roc_example}, the ROC curve of the discriminator is analyzed, and the point nearest to the top-left corner is selected as the threshold.
Discriminator outputs with AUC values equal to or exceeding this threshold are classified as belonging to the \texttt{known} class, while outputs below the threshold are classified as the \texttt{unknown} class.

\section{Evaluation}
\label{sec:eval}

We evaluated the performance of the \texttt{unknown} class discriminator across three key dimensions:
\begin{itemize}
    \item the impact of varying the ratio between \texttt{known} and \texttt{unknown} classes,
    \item the performance when encountering new \texttt{unknown} gesture classes, and
    \item the performance in cross-domain scenarios.
\end{itemize}

Three metrics are employed to assess the model's performance: Activation Action Error Rate (\textbf{AER}), Classification Accuracy (\textbf{ACC}), and Accuracy Recovery Rate (\textbf{ARR}).
\textbf{AER} represents the proportion of incorrect execution actions relative to all output actions, with actions determined based on gestures classified by the trained model.
\textbf{ACC} denotes the proportion of correct execution actions after rejecting \texttt{unknown} classes, calculated as $ACC = 1 - AER$.
\textbf{ARR} evaluates the ratio of the model's accuracy to the accuracy obtained on the training set containing only \texttt{known} classes.
Theoretically, after rejecting \texttt{unknown} classes, the model's accuracy should converge to the accuracy of a dataset comprising only \texttt{known} classes.
However, this accuracy may vary depending on the ratio of \texttt{known} to \texttt{unknown} classes and the introduction of new \texttt{unknown} gesture classes.

\subsection{Impact of Known-to-Unknown Ratio}
\label{sec:ratio}

The objective of the trained \texttt{unknown} class discriminator is to identify the \texttt{unknown} class and reject the execution of the corresponding action.
The performance of the discriminator, however, is influenced by the ratio of \texttt{known} to \texttt{unknown} classes used during training.
To evaluate the impact of varying ratios, two evaluation setups are employed.

\begin{figure}[t]
	\centering
	\includegraphics[width=\columnwidth]{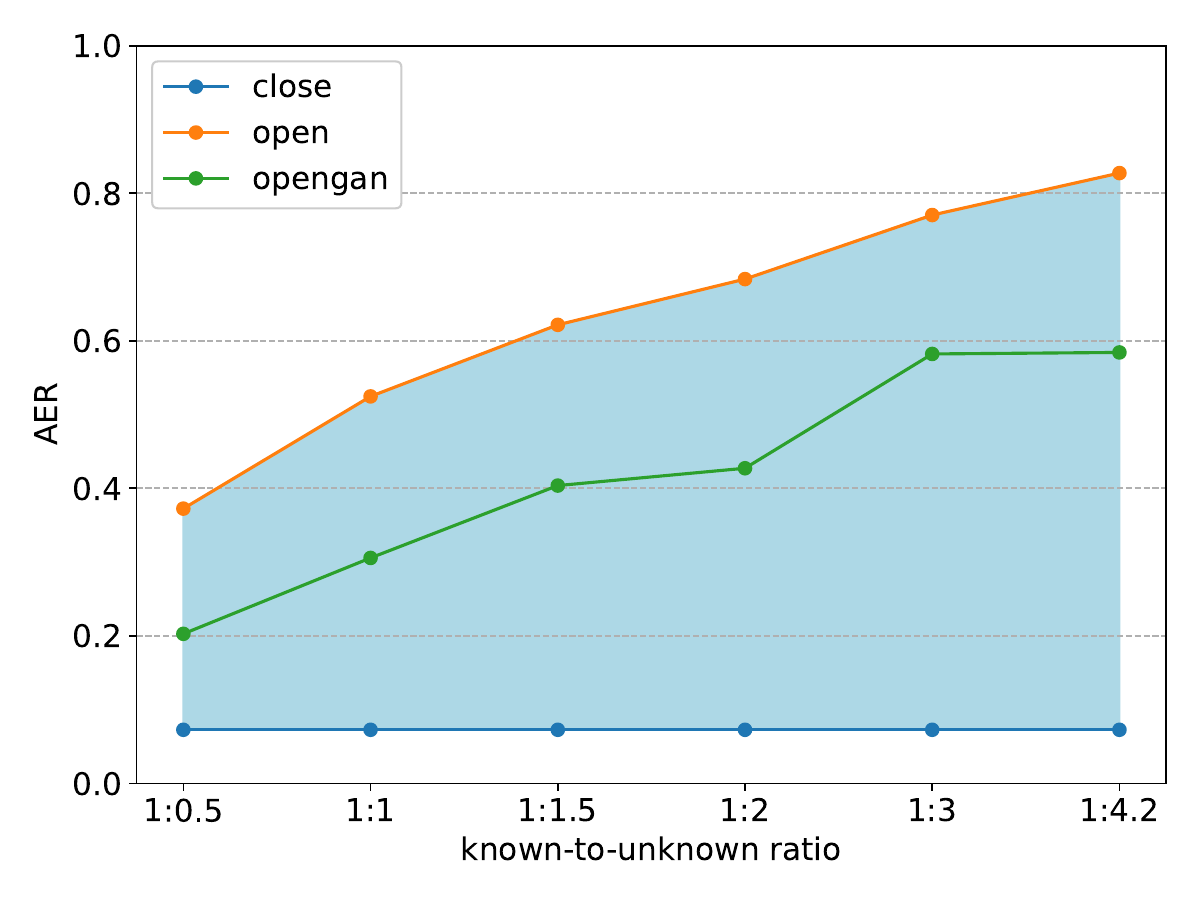}
	\caption{
            The average AER comparison of the discriminators using 10 \texttt{known} classes and 5, 10, 15, 20, 30, and 42 \texttt{unknown} classes for training.
        }
	\label{fig:exp_unknown_change}
\end{figure}

In the first setup, 10 gesture samples from the DB1 dataset are selected as \texttt{known} classes.
From the remaining dataset, 5, 10, 15, 20, 30, and 42 gesture samples are selected as \texttt{unknown} classes.
This configuration creates six discriminators trained with different ratios of \texttt{known} to \texttt{unknown} classes, ranging from $1:0.5$ to $1:4.2$.
Fig.~\ref{fig:exp_unknown_change} presents the AER values obtained using the proposed \textit{OpenGAN} approach, compared to the performance of a method relying solely on the classifier trained with the same dataset, referred to as \textit{Open} in the figure.
To establish a baseline, we also compare with a classifier trained on the entire DB1 dataset, assuming all samples are \texttt{known}.
This baseline approach, termed \textit{Close}, is not practical in real-world scenarios since \texttt{unknown} samples that the classifier has not been trained on will inevitably exist.

As illustrated in Fig.~\ref{fig:exp_unknown_change}, the AER increases as the number of \texttt{unknown} classes grows for both the \textit{OpenGAN} and \textit{Open} approaches.
However, the AER of the \textit{OpenGAN} approach consistently remains lower than that of the \textit{Open} approach.
This improvement arises from the \textit{OpenGAN} method's ability to utilize the discriminator to prevent \texttt{unknown} classes from being misclassified as \texttt{known} classes.

\begin{figure}[!t]
	\centering
	\includegraphics[width=\columnwidth]{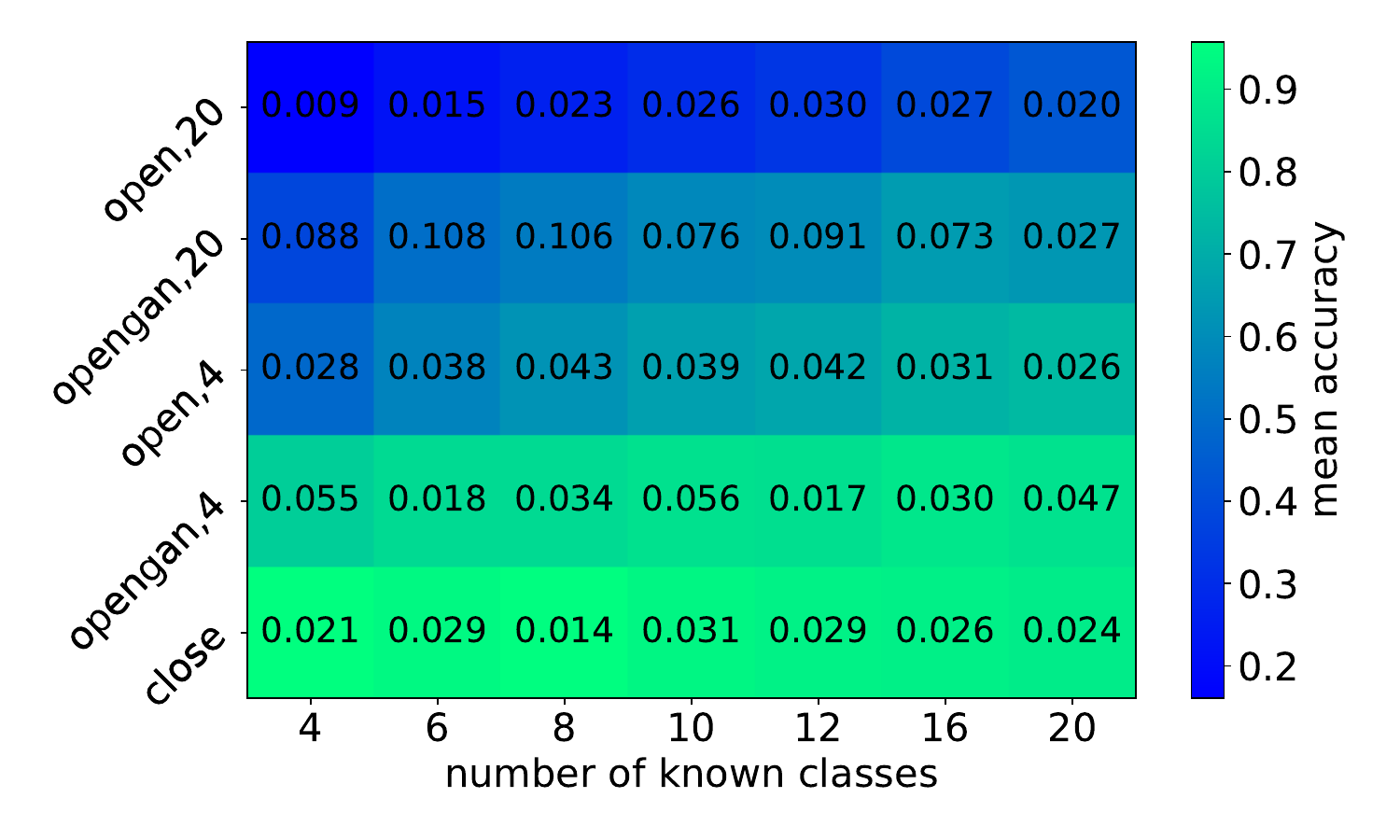}
	\caption{
        The mean accuracy heatmap of models using 4 and 20 \texttt{unknown} classes alongside 4, 6, 8, 10, 12, 16, and 20 \texttt{known} classes for training and validation.
        The number over on each cell inidcates the standard variance among 5 different subjects.
    }
	\label{fig:exp_known_change}
\end{figure}

In the second setup, we select 4 and 20 gesture samples from the DB1 dataset as \texttt{unknown} classes.
From the remaining dataset, we select 4, 6, 8, 10, 12, 16, and 20 gesture samples as \texttt{known} classes.
This setup results in 14 different ratios, with 7 ratios for each case of 4 and 20 \texttt{unknown} classes.
Fig.~\ref{fig:exp_known_change} presents the accuracies corresponding to each ratio.
As the number of \texttt{known} classes increases, the proportion of the dataset available for training also increases, leading to improved model accuracy, as shown in the figure.
For the same reason, the accuracies in the case with 20 \texttt{unknown} classes are lower compared to the case with only 4 \texttt{unknown} classes.
The proposed \textit{OpenGAN} approach with 4 \texttt{unknown} classes demonstrates the performance closest to the baseline \textit{Close} case, particularly when the number of \texttt{known} classes is large.
Compared to the \textit{Open} approach, the \textit{OpenGAN} accuracy with 4 \texttt{unknown} classes is improved by 50\% to 70\%, maintaining an accuracy range of approximately 80\% to 90\%.

Overall, the proposed \textit{OpenGAN} approach, incorporating an \texttt{unknown} class discriminator, demonstrates significant improvement compared to the \textit{Open} approach.
This improvement is consistent across various ratio settings between the numbers of \texttt{known} and \texttt{unknown} classes.
A higher ratio of \texttt{known} to \texttt{unknown} classes leads to better performance for \textit{OpenGAN}.
Notably, when the ratio exceeds $1:1$, the performance of \textit{OpenGAN} approaches the baseline, represented by the \textit{Close} approach.

For the experiment evaluating the impact of the \texttt{known}-to-\texttt{unknown} ratio, data from 5 subjects in the dataset are utilized.
Each subject is used to train a \texttt{known} class classifier.
These classifiers predict all \texttt{known} and \texttt{unknown} class data to extract feature values, which are subsequently used to create a new dataset for training the GAN and selecting the discriminator.
Tab.~\ref{tab:auc} presents the AUC values of the \texttt{unknown} class discriminators across different \texttt{known}-to-\texttt{unknown} ratios for each subject.
The results indicate that the AUC values remain relatively stable regardless of the number of \texttt{unknown} classes.
Tab.~\ref{tab:f1_score} displays the F1-scores of the discriminators, showing a downward trend as the number of \texttt{unknown} classes increases.
This metric is computed only after determining the optimal threshold from the ROC curve and, therefore, cannot serve as an indicator during the training process.

\begin{table}[!t]
\begin{center}
	\centering
	\renewcommand{\arraystretch}{1.3}
	\label{tab:auc}
	\caption{The unknown discriminator's AUC on test dataset}
	\begin{tabular}{|c|c|c|c|c|c|}
		\hline
		\textbf{Kn/Un} & \textbf{Sub 1} & \textbf{Sub 2} & \textbf{Sub 3} & \textbf{Sub 4} & \textbf{Sub 5} \\
		\hline
		1:0.5 & 0.8567 & 0.7330 & 0.7529 & 0.8291 & 0.7804 \\
		\hline 
		1:1   & 0.8100 & 0.7350 & 0.7658 & 0.8367 & 0.7215 \\
		\hline 
		1:1.5 & 0.7904 & 0.7310 & 0.7360 & 0.8709 & 0.7328 \\
		\hline 
		1:2   & 0.7801 & 0.7369 & 0.7291 & 0.8355 & 0.7570 \\
		\hline 
		1:3   & 0.7908 & 0.7839 & 0.7107 & 0.8561 & 0.7788 \\
		\hline 
		1:4   & 0.8189 & 0.7850 & 0.7719 & 0.8799 & 0.7614 \\
		\hline
	\end{tabular}%
	\end{center}
\end{table}%

\begin{table}[!t]
	\begin{center}
		\centering
		\renewcommand{\arraystretch}{1.3}
		\label{tab:f1_score}
	\caption{The unknown discriminator's F1-score on test dataset}
	\begin{tabular}{|c|c|c|c|c|c|}
		\hline
		\multicolumn{1}{|l|}{\textbf{Kn/Un}} & \textbf{Sub 1} & \textbf{Sub 2} & \textbf{Sub 3} & \textbf{Sub 4} & \textbf{Sub 5} \\
		\hline
    	1:0.5 & 0.85542 & 0.78866 & 0.79887 & 0.78412 & 0.77844 \\
		\hline
		1:1   & 0.76571 & 0.69091 & 0.73864 & 0.78070 & 0.70968 \\
		\hline
		1:1.5 & 0.7040 & 0.62115 & 0.64322 & 0.76667 & 0.68132 \\
		\hline
		1:2   & 0.62353 & 0.60045 & 0.62557 & 0.67574 & 0.62151 \\
		\hline
		1:3   & 0.55106 & 0.57971 & 0.48328 & 0.66801 & 0.56954 \\
		\hline
		1:4   & 0.54108 & 0.48130 & 0.68578 & 0.62028 & 0.53526 \\
		\hline
	\end{tabular}%
\end{center}
\end{table}%

\subsection{Performance on Different Numbers of Unknown Classes}
\label{sec:diff_number}

The \texttt{known} classes are typically predefined, either through laboratory experiments or real-world applications.
In contrast, the number of \texttt{unknown} classes is unpredictable.
When a myoelectric control system is deployed, additional \texttt{unknown} classes often emerge beyond initial expectations.
These unforeseen classes are not included in the validation set used to evaluate the discriminator.
This section examines the performance of the \textit{OpenGAN} approach under varying numbers of \texttt{unknown} classes in the validation set.

\begin{figure}[!t]
	\centering
	\includegraphics[width=\columnwidth]{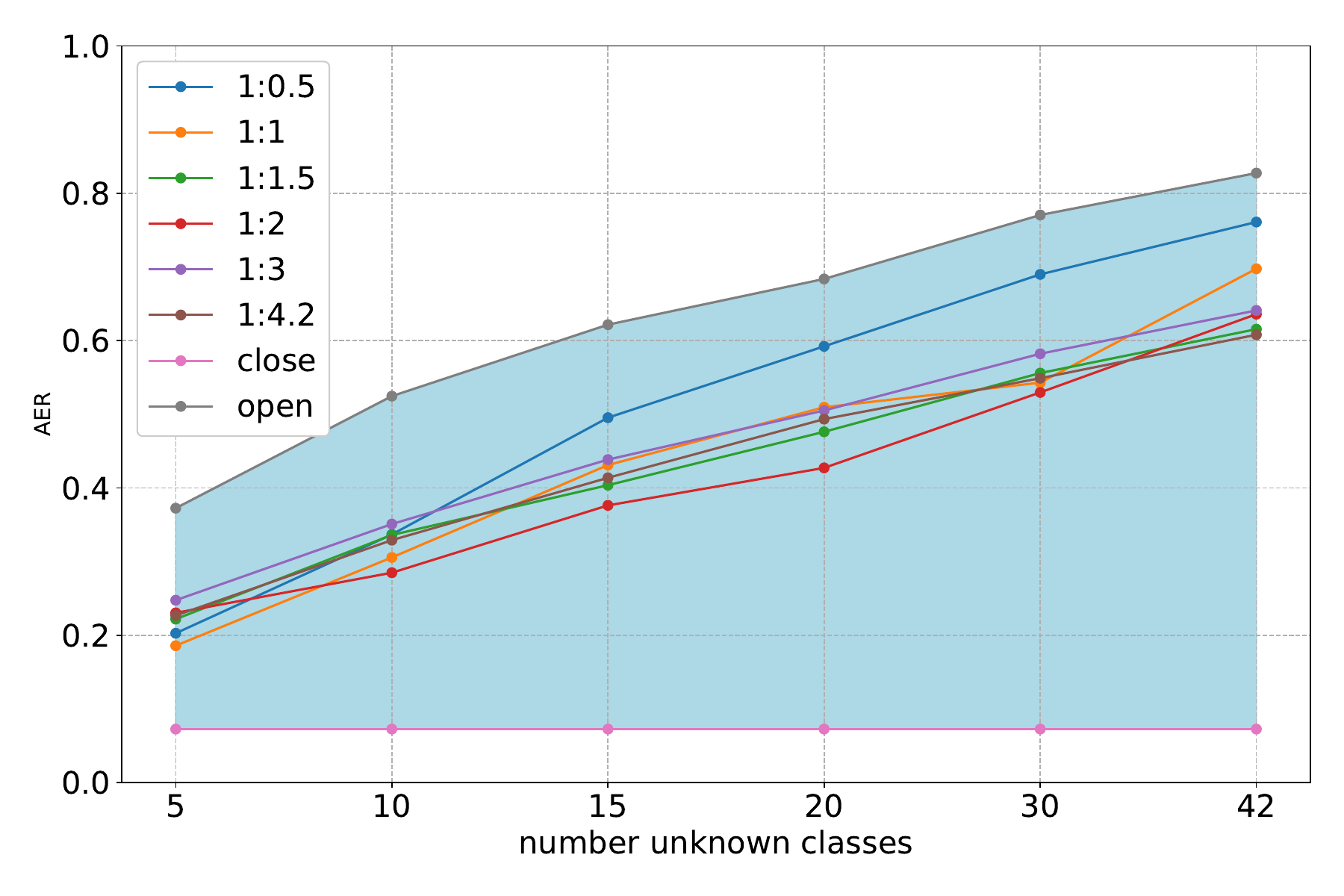}
	\caption{
            Cross Comparison of the average AER of 6 different discriminators over 6 different different \texttt{known} to \texttt{unknown} ratio of validating sets.
            The number of \texttt{known} classes is fixed to 10.
        }
	\label{fig:exp_diff_unknown}
\end{figure}

Sec.~\ref{sec:ratio} presents discriminators trained with six different ratios of \texttt{known} to \texttt{unknown} classes in the validation set.
Fig.~\ref{fig:exp_diff_unknown} illustrates the performance of these trained discriminators across various compositions of \texttt{known} and \texttt{unknown} classes in the validation set.
The number of \texttt{known} classes is fixed at 10.
The performance of the \textit{Open} and \textit{Close} methods is also evaluated, serving as the upper and lower bounds for comparison.

All six ratio discriminators demonstrate improved performance compared to the \textit{Open} method.
This suggests that the proposed discriminator method exhibits strong generalization capabilities.
Additionally, the AER values of the discriminators increase as the number of \texttt{unknown} classes in the validation set grows.
The discriminator trained with a 1:2 ratio out-performs over others, achieving the lowest AER in the validation set composition with 10, 15, 20 and 30 
\texttt{unknown} classes.
In contrast, the discriminator trained with a 1:0.5 ratio shows the worst performance when the validation set contains more than 10 \texttt{unknown} classes.

Optimal performance is not necessarily achieved by training the model using the corresponding ratio of \texttt{known} to \texttt{unknown} classes. 
As the percentage of \texttt{unknown} classes increases significantly, the outcomes become less meaningful, with no AER values dropping below 50\% when the ratio surpasses 1:2. 
Consequently, it is practical to choose a number of \texttt{unknown} actions that closely matches the number of \texttt{known} classes for forming a test set to obtain the best outcomes.

\subsection{Experiment of Cross Domain Scenario}

The deployment of a classifier, which has been trained on EMG signal samples obtained from a single individual, on samples from different individuals constitutes a cross-domain scenario. 
In order to evaluate the cross-domain effectiveness of our methodology, discriminators were trained on samples from one subject and subsequently tested on different subjects. 
These discriminators were trained utilizing six distinct \texttt{known} to \texttt{unknown} ratios, with a consistent incorporation of 10 fixed \texttt{known} classes as described in Sec.~\ref{sec:ratio}. 
Our approach does not integrate transfer learning or confidence estimation mechanisms to enhance cross-domain performance.

\begin{figure}[!t]
	\centering
	\includegraphics[width=\columnwidth]{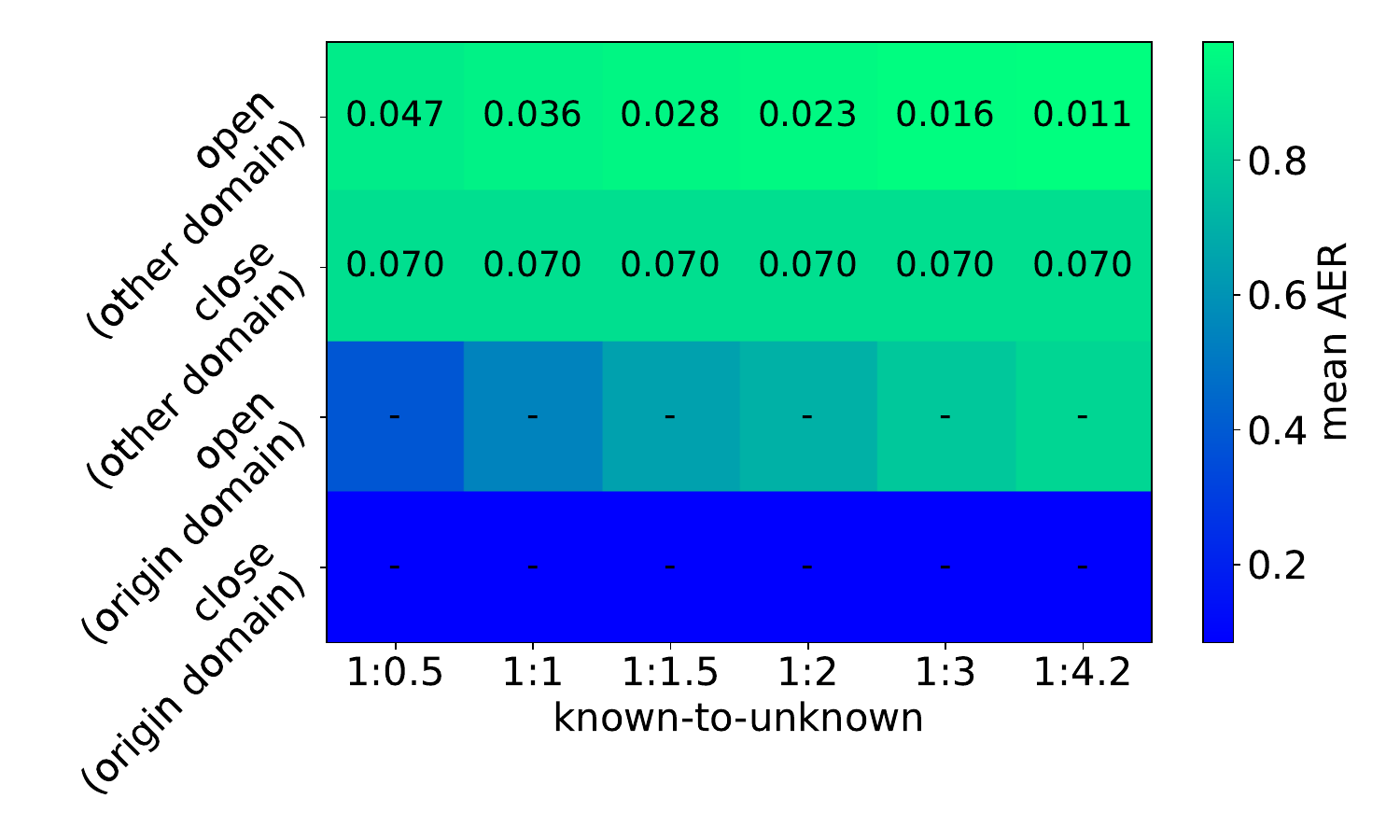}
	\caption{
          The AER of discriminators within the original training domain and with other test domains.
            \textit{close (original domain)} and \textit{open (original domain)} refer to the closed set and open set test outcomes within the original domain.
            \textit{Close (other domain)} and \textit{open (other domain)} indicate the closed set and open set results in 4 other different domains.
           The standard deviations are shown above the relevant cells.
        }
	\label{fig:exp_cross_domain_origin}
\end{figure}

As shown in Fig.~\ref{fig:exp_cross_domain_origin}, the cross-domain test results exhibit significantly higher AER values compared to those obtained from the original subject sample set.
Regardless of whether the \textit{Close} or \textit{Open} methods are used, the AER values in the cross-domain context consistently exceed 80\%.

\begin{figure}[!t]
\centering
\includegraphics[width=\columnwidth]{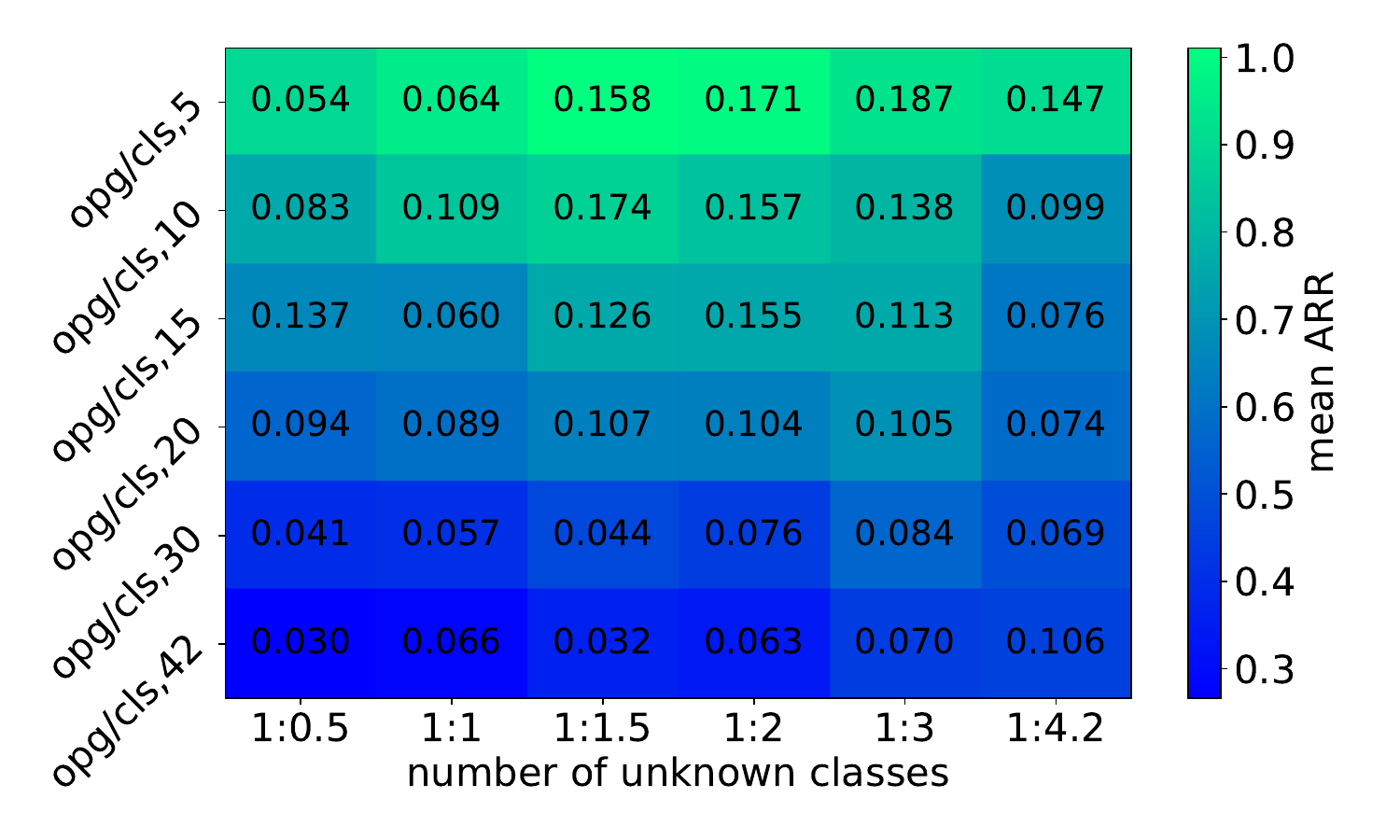}
\caption{
    The accuracy recovery rate (ARR) between \textit{OpenGAN} and \textit{Close}, represented as \textit{''opg/cls''}.
    The heatmap shows the mean ARR values of the cross-domain evaluation over 4 different subjects.
    The standard deviations are shown above the relevant cells.
}
\label{fig:exp_cross_domain_gan}
\end{figure}

To evaluate cross-domain performance between the \textit{OpenGAN} and \textit{Close} approaches, we utilized the Accuracy Recovery Rate (ARR) metric.
ARR quantifies the accuracy ratio between \textit{OpenGAN} and \textit{Close}, with a higher ARR indicating better cross-domain performance of the \textit{OpenGAN} approach.
We analyzed the ARR performance of six ratio discriminators across various \texttt{unknown} class validation sets, with the results presented in Fig.~\ref{fig:exp_cross_domain_gan}.

For the validation set with 5 \texttt{unknown} classes, referred to as the \textit{opg/cls,5} category, all ARR values exceed 90\%.
For the validation set with 30 \texttt{unknown} classes, referred to as the \textit{opg/cls,30} category, the ARR values for all models are below 50\%, except for the model trained with a 1:3 ratio of \texttt{known} to \texttt{unknown} classes.
The performance is even worse in the \textit{opg/cls,42} category.
These results indicate that the performance of the trained models decreases as the number of \texttt{unknown} classes in the validation set increases, which is an expected trend.

\begin{figure*}[!t]
	\centering
	\includegraphics[width=\textwidth]{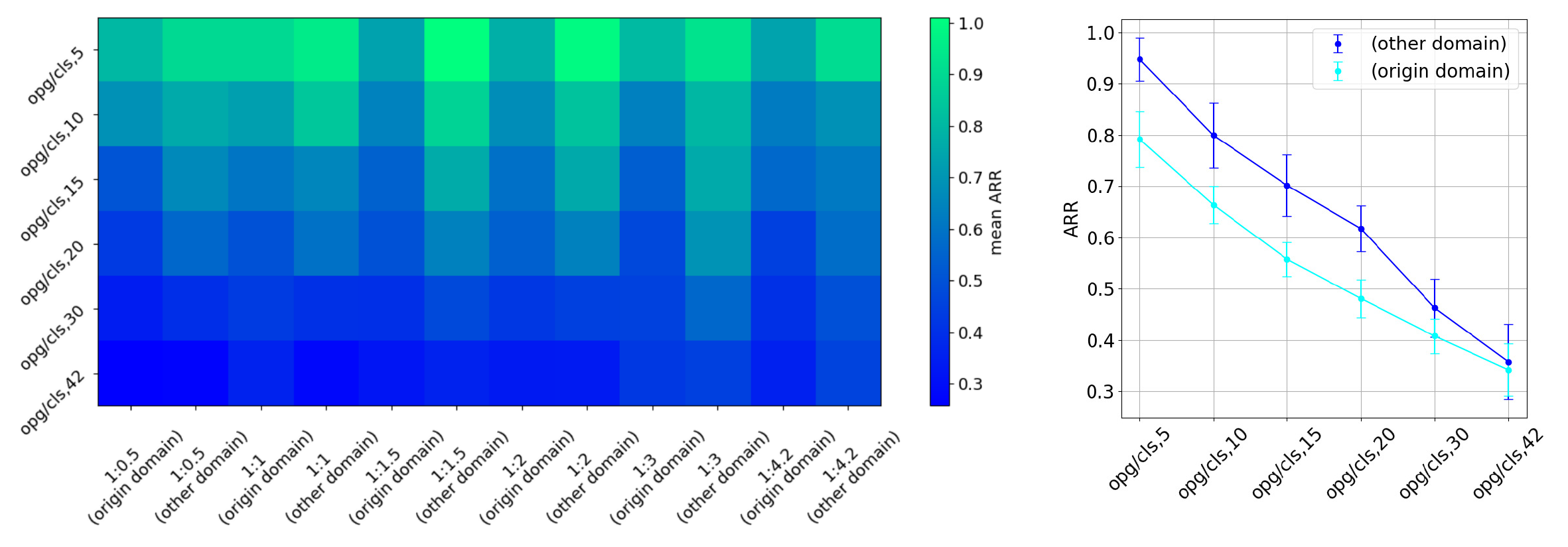}
	\caption{
        The right heatmap plot shows the mean ARR values of models trained using 6 different ratios of \texttt{known} to \texttt{unknown} and tested under cross-domain, i.e. (other domain), and original domain.
        The testing/validating set with different numbers of \texttt{unknown} classes are applied during the test.
        The right error plot indicates the improvement of \textit{OpenGAN} over \textit{Close} set based models (ARR) under cross-domain case, is higher than orginal domain. 
    }
	\label{fig:exp_cross_domain_arr}
\end{figure*}

Fig.~\ref{fig:exp_cross_domain_arr} compared cross-domain (i.e. Fig.~\ref{fig:exp_cross_domain_gan}) and original domain performance.
To the heatmap plot on the left side, the models trained with different ratios of \texttt{known} to \texttt{unknown} dataset have similar ARR values, regardless the number of \texttt{unknown} class data in the validation set.
However, the ARRs of cross-domain cases, i.e. marked as other domain, exhibit a better performance comparing to origin domain, meaning tested using the same subject.
The error plot on the right side indicated this conclusion as well.
It calculated the average ARR values over different number of \texttt{unknown} classes dataset.
The ARR of cross-domain cases are always above the original domain.
Since the ARR represents the accuracy recovery rate, which focuses on the improvement, this result the improvement of \textit{OpenGAN} based model over \textit{Close} set based mode under cross-domain, is higher than original domain testing.

\subsection{Verification of Self-Collected EMG signal}
\label{sec:verify_self}

\begin{figure}[!t]
	\centering
	\includegraphics[width=\columnwidth]{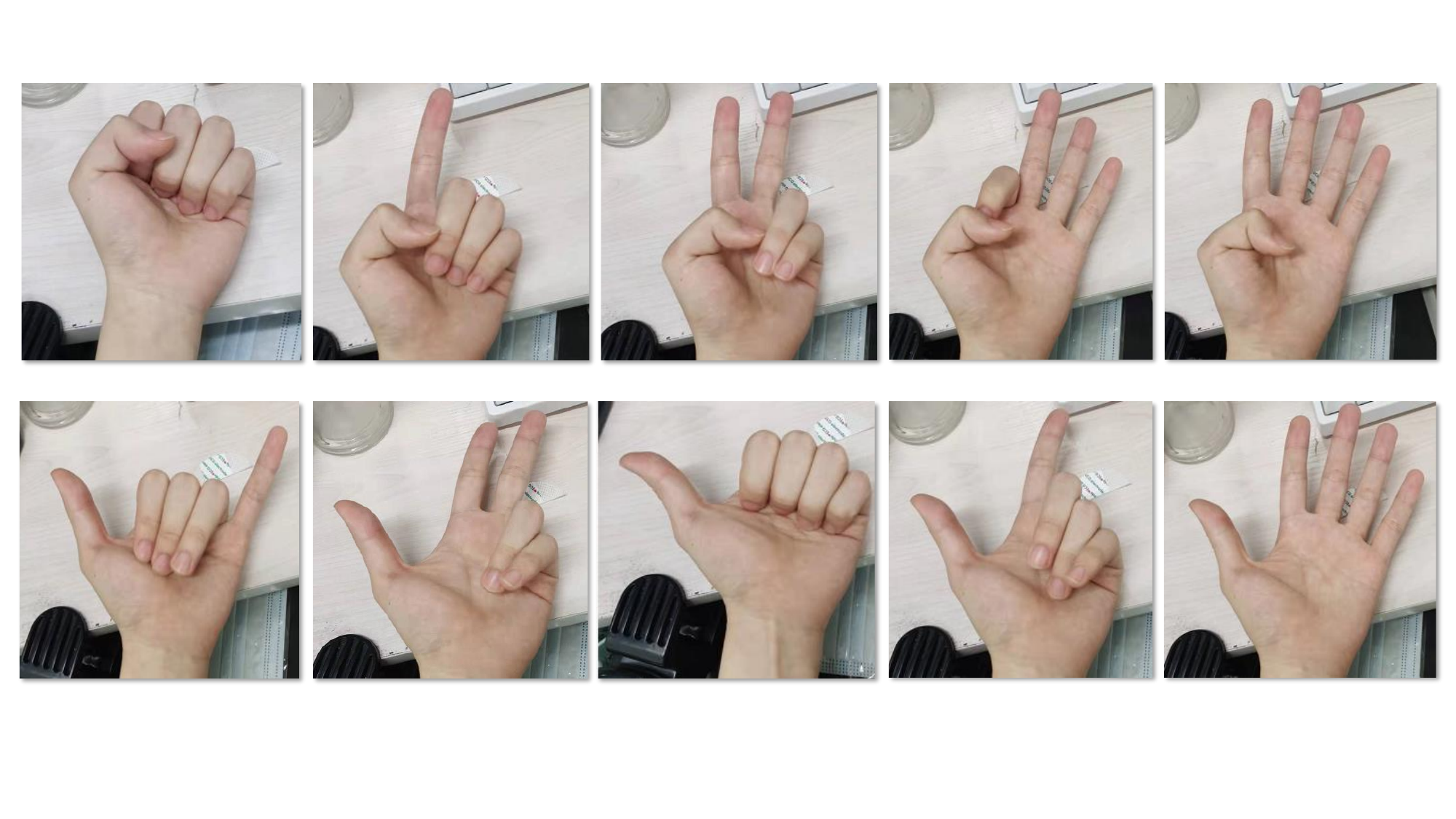}
	\caption{10 self-collected gestures for testing.}
	\label{fig:collected_gestures}
\end{figure}

In addition to training on the Ninapro DB1 dataset, we used an acquisition device to collect EMG signals from 10 different gestures, as shown in Fig.~\ref{fig:collected_gestures}.
Six of the gestures were selected as \texttt{known} classes for training the classification model, while the remaining four were designated as \texttt{unknown} classes to form the validation set.
The model achieved 97.6\% accuracy on the \texttt{known} class dataset.
The \texttt{unknown} class discriminator was trained using a GAN-based approach.

The trained CNN achieved 57.6\% open set accuracy, which improved to 81.2\% after rejecting \texttt{unknown} classes with the discriminator.
We compared the performance of the discriminator with the results in Sec.~\ref{sec:ratio}, where it was trained with 4 \texttt{unknown} classes and 6 \texttt{known} classes.
By rejecting \texttt{unknown} classes, our approach significantly reduces the error rate of the actual output actions, thereby enhancing the overall usability of the system.

\section{Discussion}
\label{sec:discussion}

We have demonstrated that the proposed GAN-based open-set recognition method effectively addresses the challenges of identifying unknown gestures in myoelectric control systems.
The discussion in this section explores how integrating this method with existing techniques, such as confidence-based rejection and transfer learning, could further enhance system stability and adaptability, addressing gaps observed in the experimental results.

\subsection{Confidence-Based Rejection Integration}

Confidence-based rejection has proven effective in reducing the Active Error Rate (AER) by ensuring that only high-confidence classifications are executed.
Our results demonstrate that the GAN-based discriminator significantly reduces AER by rejecting \texttt{unknown} gestures, particularly when the \texttt{known}-to-\texttt{unknown} ratio is favorable (less than 1:2).
However, our method does not explicitly evaluate the confidence level of predictions for \texttt{known} gestures.
Combining our approach with confidence-based rejection could further improve performance by addressing both \texttt{unknown} gesture rejection and low-confidence errors within \texttt{known} gestures.
For example, the discriminator's high \textit{AUC} and \textit{F1-scores} in known-to-unknown ratio experiments suggest that it could complement confidence-based rejection strategies to improve decision reliability across diverse scenarios.

\subsection{Enhancing Cross-Domain Generalization with Transfer Learning}

Our experiments revealed limitations in cross-domain performance, with significantly higher AER values when applying the model to new subjects.
While our method excels in distinguishing between \texttt{known} and \texttt{unknown} actions within the training domain, transfer learning could enhance its generalization to new users and sessions.
By adapting the classifier to multiple domains, transfer learning would increase the accuracy of \texttt{known} gesture classification, particularly for datasets with diverse subjects, as seen in the cross-domain experiments.
Incorporating this technique could also raise the upper limit of the \textit{accuracy recovery rate (ARR)} observed in our evaluations, making the system more robust in real-world applications.

\subsection{Comprehensive System Improvement for Real-World Deployment}

The simplicity and low computational requirements of the proposed GAN-based method make it highly suitable for deployment on resource-constrained edge devices such as prosthetics.
The model achieved significant improvements in rejecting \texttt{unknown} gestures (e.g., a 23.6\% reduction in active errors) while maintaining high accuracy (97.6\%) on {known} classes.
These findings confirm the method's potential as an effective pre-execution safeguard in open-set environments.
Combining it with confidence-based rejection and transfer learning would enable a comprehensive framework, improving classification accuracy, adaptability to new users, and error mitigation in \texttt{unknown} scenarios.
This could address gaps identified in our cross-domain tests and enhance system usability in practical settings.

\section{Conclusion}
\label{sec:conc}

This paper presents a novel GAN-based open-set recognition approach to enhance the robustness and usability of myoelectric control systems.
The method effectively rejects unknown gestures while maintaining accurate execution of \texttt{known} actions, achieving a significant improvement in both system stability and Active Error Rate (AER).
Experimental results demonstrated a high accuracy of 97.6\% for \texttt{known} classes and a 23.6\% reduction in AER after rejecting \texttt{unknown} gestures, underscoring the method's effectiveness.

The proposed approach is lightweight and computationally efficient, making it suitable for deployment on edge devices such as prosthetics.
Its ability to operate without prior exposure to unknown gestures highlights its adaptability and generalization capabilities, especially under inter-subject and inter-session variability.
However, challenges in cross-domain performance and robustness improvements for \texttt{known} gestures remain, presenting opportunities for further refinement.

Future work will explore integrating this method with confidence-based rejection to improve decision-making reliability for \texttt{known} gestures.
Additionally, the application of transfer learning could enhance cross-domain adaptability, addressing the limitations observed in current cross-subject evaluations.
Expanding the method to include applications like angle and moment estimation could further broaden its impact in bioelectric signal processing.

In conclusion, the GAN-based open-set recognition framework represents a practical and scalable solution for improving the stability, accuracy, and usability of myoelectric control systems.
Its integration with complementary techniques offers the potential to set a new standard for robustness in open-environment bioelectric control systems.

\section*{Acknowledgment}

This work is funded by
    the Guangdong Provincial Key Lab of Integrated Communication, Sensing and Computation for Ubiquitous Internet of Things, with Grant numbers 2023B1212010007 and
    Guangzhou Municipal Science and Technology Project 2023A03J0011.

\bibliographystyle{IEEEtran}
\bibliography{wang24gan}

\vfill

\end{document}